\documentclass[10pt,twocolumn,letterpaper]{article}

\usepackage{cvpr}              
\usepackage{bbm}
\usepackage{bbding}
\usepackage{booktabs, multirow, multicol}
\usepackage{color, xcolor}
\usepackage[table]{xcolor}
\usepackage{pifont}
\usepackage{makecell}
\definecolor{cvprblue}{rgb}{0.21,0.49,0.74}
\usepackage[pagebackref,breaklinks,colorlinks,allcolors=cvprblue]{hyperref}

\def\paperID{8601} 
\def\confName{CVPR}
\def\confYear{2026}

\title{OMG-Bench: A New Challenging Benchmark for Skeleton-based Online Micro Hand Gesture Recognition}

\author{Haochen Chang$^{1}$, \; Pengfei Ren$^{2}$\thanks{Corresponding Author.}, \; Buyuan Zhang$^{3}$, \; Da Li$^{4}$, \; Tianhao Han$^{3}$, \; Haoyang Zhang$^{5,6}$,\\ \; Liang Xie$^{5,6}$, \; Hongbo Chen$^{1}$, \; Erwei Yin$^{5,6}$\footnotemark[1]\\
$^{1}$School of Systems Science and Engineering, Sun Yat-sen University\\
$^{2}$Beijing University of Posts and Telecommunications \quad 
$^{3}$Shanghai Jiao Tong University \\
$^{4}$Nankai University \quad
$^{5}$Defense Innovation Institute, Academy of Military Sciences\\
$^{6}$Tianjin Artificial Intelligence Innovation Center\\
{\tt\small changhch5@mail2.sysu.edu.cn, rpf@bupt.edu.cn, yinerwei1985@gmail.com}\\
\url{https://omg-bench.github.io/}
}

\begin{document}
\maketitle
\begin{abstract}
Online micro gesture recognition from hand skeletons is critical for VR/AR interaction but faces challenges due to limited public datasets and task-specific algorithms. Micro gestures involve subtle motion patterns, which make constructing datasets with precise skeletons and frame-level annotations difficult. To this end, we develop a multi-view self-supervised pipeline to automatically generate skeleton data, complemented by heuristic rules and expert refinement for semi-automatic annotation. Based on this pipeline, we introduce \textbf{OMG-Bench}, the first large-scale public benchmark for skeleton-based online micro gesture recognition. It features 40 fine-grained gesture classes with 13,948 instances across 1,272 sequences, characterized by subtle motions, rapid dynamics, and continuous execution. To tackle these challenges, we propose \textbf{Hierarchical Memory-Augmented Transformer (HMATr)}, an end-to-end framework that unifies gesture detection and classification by leveraging hierarchical memory banks which store frame-level details and window-level semantics to preserve historical context. In addition, it employs learnable position-aware queries initialized from the memory to implicitly encode gesture positions and semantics. Experiments show that HMATr outperforms state-of-the-art methods by 7.6\% in detection rate, establishing a strong baseline for online micro gesture recognition. 
\end{abstract}    
\section{Introduction}
\label{sec:intro}
Recent advances in hand pose estimation ~\cite{han2022umetrack, ren2025prior, ren2023decoupled, ren2023two, ren2025rule, liu2024keypoint} enable head-mounted displays such as Meta Quest and PICO to capture hand skeletons easily. Skeleton data, robust to viewpoint and illumination changes, allow skeleton-based gesture recognition ~\cite{liu2020decoupled, de2016skeleton, liu2023temporal} to perform well in VR/AR interaction, attracting significant research interest.

\begin{figure}[t]
	\centering
	\includegraphics[width=\linewidth]{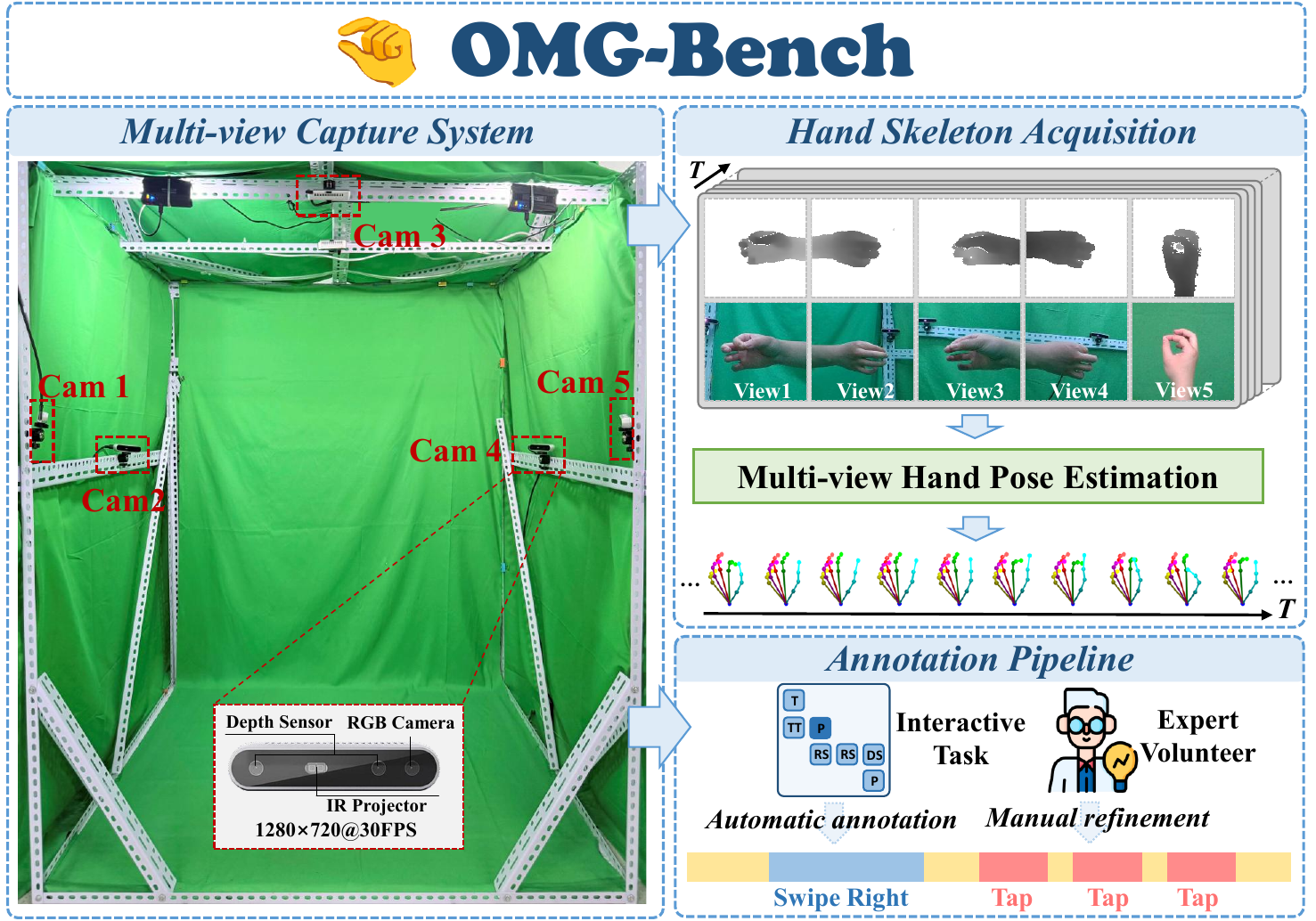}
	\caption{Data collection and annotation pipeline of OMG-Bench, using a calibrated five-camera RGB-D system and self-supervised multi-view hand pose estimation to obtain high-quality skeletons, followed by semi-automatic frame-level gesture labeling.}
	\label{dataset}
\end{figure}

In real-world scenarios, gesture recognition algorithms need to process skeleton data in a streaming and online manner, which requires datasets to reflect realistic, continuous interaction dynamics. However, existing skeleton-based online gesture recognition datasets suffer from several limitations: \textbf{(1) Scale.} The datasets are small, containing limited gesture sequences, few valid samples, and insufficient diversity in categories. \textbf{(2) Quality.} Skeleton data are captured using outdated, device-integrated single-view hand pose estimators, resulting in noisy and inaccurate joint position information that hampers algorithm training. \textbf{(3) Dynamics.} The datasets overlook the fast and continuous execution of gestures, as most samples present isolated gestures with clear temporal gaps, failing to realistically simulate scenarios where users perform multiple rapid and consecutive gestures in real interactions.

It is worth noting that the gestures in the aforementioned datasets are predominantly \textbf{conventional macro gestures with large motion amplitudes}. Prolonged interaction using such gestures often leads to user arm muscle fatigue ~\cite{wang2024beyondvision}. In contrast, micro gestures involve subtle motions and can effectively reduce physical burden and improve user experience ~\cite{chan2016user, kin2024stmg}. While micro gesture recognition based on non-visual sensors (e.g., terahertz radar ~\cite{wang2020negative}, electromyography ~\cite{wang2024beyondvision}, ultrasound sensors ~\cite{ling2020ultragesture}) has been studied, skeleton-based micro gesture recognition methods remain scarce. There is currently no publicly available skeleton-based micro gesture dataset, which significantly limits research in skeleton-based online micro gesture recognition.

To bridge these gaps, we introduce \textbf{OMG-Bench}, the first publicly available skeleton-based \textbf{O}nline \textbf{M}icro \textbf{G}esture recognition dataset to date. We design a novel data pipeline that automatically generates high-precision skeletons from a multi-view system and employs a semi-automatic process for accurate, expert-verified frame-level annotations. OMG-Bench includes the fine-grained finger-to-finger interaction gestures such as tap, double-tap, swipe, and pinch. It contains 40 gesture classes and 1,272 high-quality skeleton gesture sequences, comprising 13,948 gesture samples with frame-level class annotations. OMG-Bench provides a powerful data foundation for online micro gesture recognition and establishes a new benchmark for gesture recognition algorithms.

The key challenges of online micro gesture recognition tasks are as follows: \textbf{(1) Subtle inter-class differences}: different micro gesture categories often differ only in fine-grained motion details (e.g., thumb-tip to index fingertip click vs. thumb-tip to middle fingertip click), resulting in highly overlapping skeleton feature distributions and frequent confusion. \textbf{(2) Rapid dynamic characteristics}: each micro gesture typically lasts for a short duration with small motion amplitude and is performed swiftly, leading to sparse effective feature information and increased difficulty for real-time recognition. \textbf{(3) Significant variation in temporal lengths}: there exist notable temporal duration differences between different micro gestures and among different executions of the same gesture (e.g., clicking vs. pinching can differ by 2 to 3 times in length), and continuous execution often results in indistinct gesture boundaries.

Existing online gesture recognition methods have certain limitations in architecture design and data stream processing. Some methods rely on two-stage heuristic gesture detection algorithms ~\cite{caputo2021shrec, shen2022gesture} or separately trained gesture detection networks ~\cite{kopuklu2019real}, which hinder end-to-end optimization. In contrast, single-stage methods employ Connectionist Temporal
Classification (CTC) loss ~\cite{molchanov2016online} for online recognition, but the brief and subtle nature of micro gesture signals often leads the network to produce blank predictions. Furthermore, both methods mentioned above adopt a sliding-window-based streaming framework, which is sensitive to hyperparameters such as stride and window size. Non-overlapping sliding windows may truncate ongoing gestures and impair recognition completeness, while overlapping windows can lead to redundant computations.

Existing sliding‑window schemes suffer from limitations in capturing the rapid dynamics of micro gestures due to insufficient cross‑window contextual information, while two‑stage schemes exacerbate the spatiotemporal ambiguity in representing subtle actions as a result of task decoupling. To address these issues, we propose a unified, end-to-end framework named \textbf{H}ierarchical \textbf{M}emory-\textbf{A}ugmented \textbf{Tr}ansformer \textbf{(HMATr)} that jointly performs gesture detection and classification. HMATr leverages a hierarchical memory mechanism to preserve continuous semantic and fine-grained details across recent historical windows, thereby maintaining both global and local temporal consistency during non-overlapping sliding-window inference. Meanwhile, the learnable position-aware queries integrate temporal memory with current observations, implicitly capturing subtle temporal variations of gestures as well as discriminative semantic representations.

The main contributions are summarized as follows:
\begin{itemize}
    \item We present OMG-Bench, the first publicly available large-scale skeleton-based online micro gesture recognition dataset, featuring 40 fine-grained, easily confusable gestures with subtle motions to facilitate advancement in micro gesture recognition algorithms.
    \item We propose HMATr, a novel end-to-end baseline for online micro gesture recognition that introduces a hierarchical memory mechanism for contextual modeling and uses learnable position-aware queries to implicitly encode gesture position and semantics in a unified architecture.
    \item We benchmark multiple SOTA skeleton-based online gesture recognition methods on OMG-Bench, establishing a comprehensive benchmark for micro gesture recognition.
\end{itemize}
\begin{table*}[htb]
    \centering 
    \renewcommand{\arraystretch}{0.9}
    \caption{Comparison between open-source skeleton-based gesture recognition datasets and the proposed OMG-Bench.}
    \vspace{-2mm}  
    \begin{tabular}{lcccccccc}
        \toprule
       & \textbf{Dataset} & \textbf{Gesture}  &  \multirow{2}*{\textbf{Classes}} &  \multirow{2}*{\textbf{Subjects}} &  \multirow{2}*{\textbf{Sequences}} &  \multirow{2}*{\textbf{Instances}} & \textbf{Skeleton} &  \textbf{Action}\\
       &  \textbf{Type} & \textbf{Type} &&&&& \textbf{Annotation} & \textbf{Annotation} \\
        \midrule
        ChAirGest~\cite{ruffieux2013chairgest} & offline & Macro & 10 & 10 & - & 1200 & Kinect V2 & -\\
        DHG~\cite{de2016skeleton} & offline & Macro & 28 & 20 & - & 2800 & RealSense & -\\
        SHREC'17~\cite{de2017shrec} & offline& Macro & 28 & 27 & - & 2800 & RealSense & - \\
        \midrule
        LMDHG~\cite{boulahia2017dynamic}& online & Macro & 13 & \textbf{21} & 50 & 608 & Leap Motion & manual \\
        ODHG~\cite{de2017shrec} & online& Macro & 28 & 20 & 280 & 2800 & RealSense &  manual\\
        SHREC'19~\cite{caputo2019shrec}& online & Macro & 5 & 13 & 195 & 195 & Leap Motion & manual \\
        SHREC'21~\cite{caputo2021shrec}& online& Macro & 17 & 5 & 180 & 720 & Leap Motion & manual \\
        SHREC'22~\cite{emporio2022shrec}& online & Macro & 16 & 6 & 288 & 1152 & HoloLens 2 & manual \\
        \rowcolor{gray!20} \textbf{OMG-Bench}& \textbf{online} & \textbf{Micro} & \textbf{40} & {18} &\textbf{1272} & \textbf{13948} & \textbf{Multi-view} & \textbf{semi-auto.} \\
        \bottomrule
    \end{tabular}
    \label{tab:dataset}
\end{table*}
\section{Related Work}
\label{sec:related}
{\bf Skeleton-based Gesture Recognition Dataset.} Skeleton-based gesture recognition has attracted much attention due to its robustness under varying conditions. Existing datasets like SHREC’17~\cite{de2017shrec} and DHG~\cite{de2016skeleton}, collected with Intel RealSense, include 28 gesture categories and 2,800 samples mainly for offline recognition. SHREC’22~\cite{emporio2022shrec}, gathered via HoloLens 2, contains 288 sequences and 1,152 instances for online recognition. However, these datasets are limited in scale, quality, and diversity, as summarized in Table \ref{tab:dataset}, which hinders progress. To overcome this, we introduce OMG-Bench, a large-scale online micro gesture recognition dataset offering a more challenging benchmark.\\
{\bf Online Gesture Recognition Methods.} Research on skeleton-based action recognition~\cite{STGCN,WDCE,chang2025hierarchical} has advanced gesture recognition. Some methods~\cite{liu2023temporal,liu2020decoupled} use GCNs with TCNs or RNNs to recognize pre-segmented gestures, but they are limited to offline scenarios. Recent online methods~\cite{cunico2023oo,shen2022gesture,chae2025online} still face sensitivity to sliding window parameters, computational redundancy, and performance drops in two-stage designs, and they are not specifically tailored to the characteristics of micro-gestures. We address these issues with a streaming framework that integrates detection and classification.\\
{\bf Micro Gesture.} Most micro gesture research focuses on other modalities~\cite{wang2024beyondvision,wang2020negative}, with limited work on skeleton-based approaches. STMG~\cite{kin2024stmg} introduced the first TCN-based online method using only thumb and index finger skeletons to recognize seven common micro gestures. Its relatively small number of gesture classes and simplified input dimensions reduce the recognition difficulty. In contrast, our OMG-Bench offers 40 gesture classes with faster temporal dynamics, posing greater recognition challenges.\\
\begin{table}[t]
    \centering 
    \setlength\tabcolsep{1.2mm} 
    \renewcommand{\arraystretch}{0.9}
    \caption{Statistical Comparison with SHREC'21/22.}
    \vspace{-2mm}  
    \begin{tabular}{lcccccccc}
        \toprule
        & \textbf{SCCGP}$\uparrow$ & \textbf{MGI}$\downarrow$ & \textbf{MGD}$\downarrow$ & \textbf{NMJD}$\downarrow$\\
        \midrule
        SHREC'21~\cite{caputo2021shrec}& 0.29\% & 12.48s & 2.60s & 158.38 \\
        SHREC'22~\cite{emporio2022shrec}& 0.09\% & 2.86s & 1.19s & 128.73\\
        \rowcolor{gray!20} \textbf{OMG-Bench} & \textbf{27.60\%} & \textbf{0.22s} & \textbf{0.57s} & \textbf{8.95}\\
        \bottomrule
    \end{tabular}
    \label{tab:dataset2}
    \vspace{-3mm}
\end{table}
\vspace{-6mm}  
\section{OMG-Bench}
\label{sec:omg}
\subsection{Data Collection and Annotation}
\label{ssec:3.1}
{\bf Multi-view Capture System.} OMG-Bench dataset is collected using a multi-view synchronized capture platform configured with five Intel RealSense D415i cameras, as shown in Figure \ref{dataset}. Each camera records high-definition RGB-D video streams at a resolution of 1280×720 and 30 FPS. The system is calibrated using a 3D calibration target ~\cite{ha2017deltille}, achieving a pixel-level root mean square reprojection error of 0.42–0.48 pixels.\\
{\bf Data Collection.} Inspired by prior work~\cite{kin2024stmg}, we adopt an interactive data collection setup in which subjects perform gestures through real-time feedback tasks. To provide online triggers during data collection, we design a heuristic recognizer based on finger-joint distances and predefined rules. Since the gesture sequence is known in advance, the recognizer only needs to check the finger configuration of the current target gesture, rather than distinguish among all possible gestures or finger combinations. This simplifies online recognition and enables relatively accurate real-time triggering, making the heuristic recognizer practical for data collection. Meanwhile, loose constraints encourage natural gesture execution and reduce missed detections.

A total of 18 subjects (11 males and 7 females) participated in data collection. During collection, they placed their hands within the multi-view field of view and performed 70-80 randomly generated gesture sequences (8-16 instances each) following tasks. The gesture classes and order in each sequence were randomly assigned to avoid habitual or predictable motion patterns. Each subject contributed about 800 valid instances, ensuring diverse subjects and motions.\\
{\bf Hand Skeleton Acquisition.} We adopt the multi-view self-supervised hand pose estimation method ~\cite{ren2022dual,ren2022mining}, which leverages multi-view depth images to automatically generate 21-joint hand skeleton data. We test this method on the manually annotated held-out evaluation set, and the Mean Per Joint Position Error is 2.78 mm.\\
{\bf Action Annotation Pipeline.} We designed an efficient semi-automatic gesture annotation pipeline to label the start and end frames of micro gesture segments. A heuristic gesture recognition algorithm first generates initial annotations automatically. To ensure quality, five expert volunteers reviewed the automatic annotations. For sequences with poor initial annotations, volunteers manually refined them to correct deviations. Specifically, if the gestures performed by subjects differed from predefined sequence, predefined labels were modified to match. If initial boundaries differ significantly from the actual motion, they are manually re-annotated. More details on data collection and annotation are provided in the supplementary material. 
\begin{figure}[t]
	\centering
	\includegraphics[width=\linewidth]{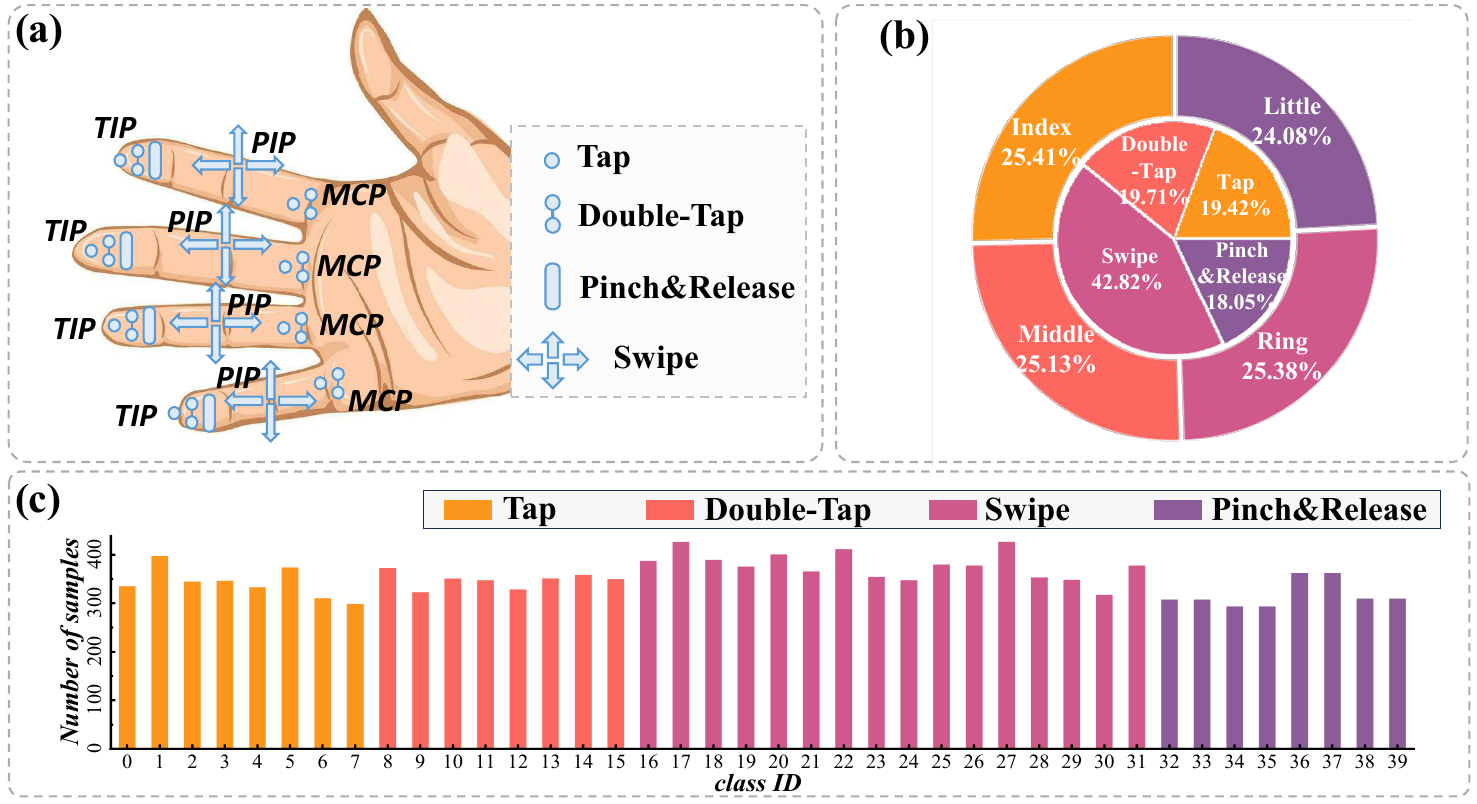}
	\caption{\textbf{Dataset properties.} \textbf{(a)} Types and locations of defined micro gestures. TIP, PIP, and MCP denote the fingertip, proximal interphalangeal joint, and metacarpophalangeal joint. \textbf{(b)} Statistics of gesture types. \textbf{(c)} Distribution of sample counts per class.}
	\label{properties}
\end{figure} 
\subsection{Dataset Statistics and Properties}
\label{ssec:3.2}
{\bf Micro Gesture Design.} We define 40 common single-hand micro gesture classes, as illustrated in Figure \ref{properties}(a). These gestures are all realized through interactions between the thumb and other fingers. The class differences are characterized by three dimensions: \textbf{(1) Interacting fingers}, i.e., the thumb interacting with the index, middle, ring, or little finger. \textbf{(2) Interaction types}, including typical actions such as single tap, double tap, slide, pinch, and release. \textbf{(3) Interaction locations}, referring to the specific contact areas between the thumb and finger—for example, taps occur at the fingertip or finger pad, while sliding covers a larger finger region. Detailed class labels and gesture descriptions are provided in the supplementary material.\\
{\bf Statistics.} The OMG-Bench dataset contains 18 participants and 1,272 micro gesture skeleton sequences recorded at 30 FPS, with a total of 13,948 micro gesture instances. The duration of each gesture instance ranges from 0.26s to 2.1s, with an average length of 0.57s. The dataset is split into training and testing sets via a cross-subject protocol: 12 subjects (897 sequences) for training and 6 subjects (375 sequences) for testing. Figure \ref{properties} presents some statistics of OMG-Bench. Table \ref{tab:dataset} compares OMG-Bench with existing open-source skeleton-based gesture recognition datasets, demonstrating that our dataset significantly surpasses others in multiple aspects. \\
{\bf Properties.} We statistically analyzed the Mean Gesture Duration (MGD), Mean Gesture Interval (MGI), Same-Class Continuous Gesture Percentage (SCCGP), and Normalized Mean Joint Displacement (NMJD) in the dataset, as shown in Table \ref{tab:dataset2}. Mean gesture duration in OMG-Bench is shorter than one second, which substantially increases recognition difficulty. Compared to SHREC’21/22 ~\cite{caputo2021shrec,emporio2022shrec}, OMG-Bench contains a higher proportion of continuous gestures of the same class and shorter mean gesture interval, which better reflect real-world scenarios but introduce ambiguity in the boundaries between gesture instances. Meanwhile, the normalized mean joint displacement in OMG-Bench is significantly smaller than in other datasets, and its subtle dynamics further exacerbate the challenges of micro gesture recognition. Overall, due to its short response times, abundant continuous same-class gestures, and subtle motion characteristics, OMG-Bench serves as a critical benchmark for evaluating online micro gesture recognition algorithms.\\
\begin{figure*}[htb]
	\centering
	\includegraphics[width=\textwidth]{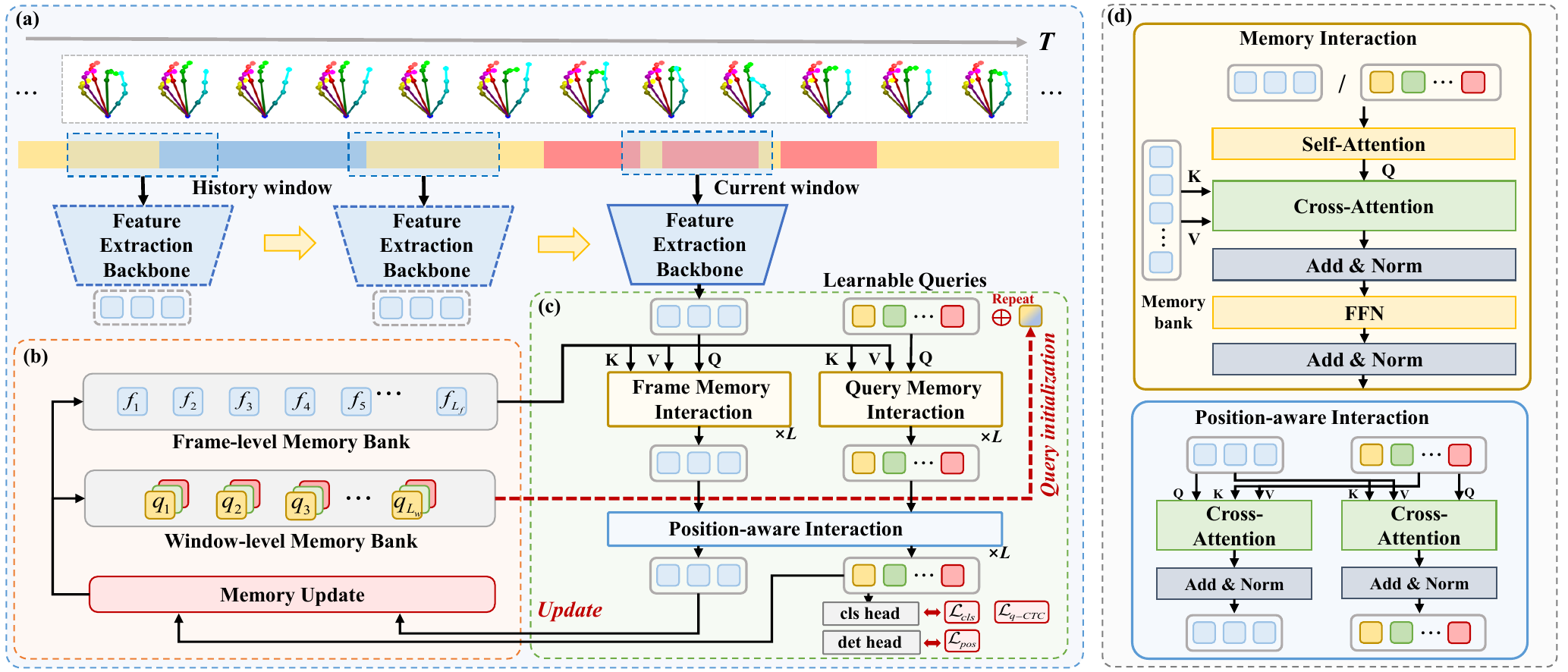}
	\caption{\textbf{Overview of our proposed HMATr}. \textbf{(a)} Lightweight backbone processes streaming skeleton inputs using a non-overlapping sliding window approach. \textbf{(b)} Hierarchical memory bank uses historical temporal information to enrich the content of the current window. \textbf{(c)} Position-aware queries implicitly capture potential hand movements, enabling unified detection and recognition. \textbf{(d)} Memory Interaction and Position-aware Interaction encode both position and semantic information of gesture instances from the memory-enhanced features.}
	\label{overview}
\end{figure*}
\vspace{-6mm}  
\section{Method}
\label{sec:method}
Online micro gesture recognition requires the model to continuously receive and process skeleton data streams $S=\{s_1, s_2, ... , s_t \}$ and output results in real time, where $s_t\in \mathbb{R}^{J \times 3}$ denotes the skeleton coordinates at the current time step $t$. Therefore, temporal modeling schemes tailored for streaming data are crucial. We propose a sliding-window streaming framework named {\bf H}ierarchical {\bf M}emory-{\bf A}ugmented {\bf Tr}ansformer \textbf{(HMATr)}, depicted in Figure \ref{overview}. The overall architecture can be divided into three parts: (1) a feature extraction backbone that embeds low-level skeleton features (Sec. \ref{ssec:4.1}), (2) a hierarchical memory bank that enriches the current window with historical temporal information (Sec. \ref{ssec:4.2}), and (3) learnable queries unifying detection and classification (Sec. \ref{ssec:4.3}). The framework is trained end-to-end with a bipartite matching loss and a query-based CTC loss (Sec. \ref{ssec:4.4}). We provide further architectural details in the following sections.

\subsection{Skeleton Feature Extraction}
\label{ssec:4.1}
The inherent topology of the skeleton can be represented as a graph composed of nodes and edges ~\cite{STGCN}. Consequently, graph convolutional network methods have gradually become the mainstream approach for skeleton feature extraction. In this work, we adopt ST-GCN ~\cite{STGCN} as our feature extraction backbone to capture local spatiotemporal features within a temporal window. Different from the original design, we modify its depth and configuration to create a more lightweight model that outputs frame-level features. 

Formally, given a skeleton sequence $S_T \in \mathbb{R}^{T_w \times J \times 3}$ containing $T_w$ frames, we feed it into the feature extraction backbone to obtain the initial skeleton embedding features $\mathbf{X}_{{embed}} \in \mathbb{R}^{B \times C \times T_w \times J}$, where $B$ is the batch size, $C$ is the output channel dimension, $T_w$ is the number of frames in the window, and $J$ is the number of joints. Next, we inject temporal positional information within the window into the frame-level features using sinusoidal positional encoding $\mathbf{PE_{win}}$, as follows:
\begin{equation}
    \mathbf{X}_{{frame}} = Reshape(GAP(\mathbf{X}_{{embed}})) + \mathbf{PE_{win}},
\label{eq:Xframe}
\end{equation}
where $\mathbf{X}_{{frame}} \in \mathbb{R}^{B \times T_w \times C}$ denotes the frame-level skeleton features extracted by the backbone, and $GAP(\cdot)$ represents global average pooling over the joint dimension.

\subsection{Hierarchical Memory Modeling}
\label{ssec:4.2}
Existing online recognition methods~\cite{cunico2023oo,caputo2021shrec} mainly rely on sliding windows, which have two main drawbacks: (1) To avoid gesture truncation, they typically use a high overlap ratio, causing computational redundancy. (2) Features are extracted independently for each window, without modeling temporal dependencies across adjacent windows. This leads to fragmented gesture understanding and misclassification of transitional gestures. Inspired by prior works~\cite{he2024ma, dang2025hallucination}, we propose a hierarchical memory bank to model cross-window temporal dependencies. It includes a frame-level memory for low-level action details and a window-level memory for high-level semantics. By leveraging multi-level historical information, this design adapts to gesture duration variations and provides disambiguation cues for confusable gestures, enabling effective online recognition with non-overlapping windows.\\
{\bf Frame-level Memory Bank.} The frame-level memory bank $\mathcal{M}_f \in \mathbb{R}^{B \times L_f \times C}$ focuses on storing frame-level skeleton features extracted from recent historical windows, where $L_f$ denotes the preset memory length. Its primary objective is to supplement the current window with critical contextual information by incorporating historical frame-level features, thereby effectively mitigating information loss caused by the non-overlapping window scheme. As a low-level skeleton feature memory, it preserves fine-grained spatiotemporal details present in the original skeleton data.\\
{\bf Window-level Memory Bank.} The window-level memory bank $\mathcal{M}_w \in \mathbb{R}^{B \times L_w \cdot N \times C}$ is dedicated to maintaining high-level semantic abstract representations within each temporal window, where $L_w$ denotes the number of historical windows maintained in memory, and $N$ represents the number of queries contained in each window. It is maintained via learnable queries bound to each window (detailed in Section \ref{ssec:4.3}), which encode the semantic and positional information of micro gestures. As a high-level memory, it transcends the fine-grained details of raw frames and focuses on capturing the holistic semantics of action sequences. This abstract representation provides task-oriented semantic cues for online micro gesture understanding, working synergistically with the frame-level memory.\\
{\bf Memory Update.} Both memory banks are maintained as fixed-length first-in-first-out queues. During each update, the memory features corresponding to the new window are appended to the queue, while the earliest features are simultaneously removed. This process can be expressed as:
\begin{equation}
\begin{split}
    &\mathcal{M}_f^t = \big[\,\mathrm{Concat}\!\left(\mathcal{M}_f^{t-1}, \mathbf{X}^t_{\mathrm{frame}}\right)\big]_{T_f-L_f+1}^{T_f}, \\
    &\mathcal{M}_w^t = \big[\,\mathrm{Concat}\!\left(\mathcal{M}_w^{t-1}, \mathbf{X}^t_{\mathrm{query}}\right)\big]_{T_w-L_w+1}^{T_w},
\end{split}
\label{eq:update_math}
\end{equation}
where $\mathcal{M}_f^t$ and $\mathcal{M}_w^t$ denote the memory banks for the $t$-th window, while $\mathbf{X}^t_{frame}$ and $\mathbf{X}^t_{query}$ represent the frame features and query features after interaction within the current window, respectively.

\subsection{Position-aware Queries}
\label{ssec:4.3}
To precisely localize the timing of gesture occurrences, some methods use a two-stage ``detect-then-classify'' scheme. However, this approach prevents end-to-end optimization between the detection and classification networks, causing the overall recognition performance to heavily depend on the detection accuracy. Furthermore, the hand skeleton remains continuously visible in the field of view, forcing the detection network to rely solely on subtle skeleton movements to decide whether a micro gesture has occurred. This intrinsic ambiguity significantly increases the task difficulty and often leads to missed or false detections. 

In fact, the occurrence timing of gestures is strongly correlated with action categories. Localization requires category cues to distinguish valid gestures from meaningless motions, while classification depends on precise temporal localization to capture key action segments. Motivated by query designs in prior detection frameworks~\cite{carion2020end,liu2022dab}, we use learnable queries $\mathbf{Q}$ as a unified medium for detection and classification. These queries interact with current and historical features to extract gesture information, capturing high-level spatial and category semantics.\\
{\bf Query Initialization.} We expect the queries to possess the ability to perceive latent motion, thereby improving gesture classification and localization. To this end, we initialize the queries using a window-level memory bank $\mathcal{M}_w^t$, which stores the historical query features. We average $\mathcal{M}_w^t$ along the temporal dimension to obtain a global memory query $\mathbf{Q}^t_m \in \mathbb{R}^{1 \times C}$, which is then added to the current window query $\mathbf{Q}^t_c \in \mathbb{R}^{N \times C}$. This can be formulated as:
\begin{equation}
    \mathbf{Q}^t_m = \frac{1}{L_w} \sum_{i=1}^{L_w} \mathcal{M}_{w,i}^{t}, \quad \mathbf{Q}^t_c = \mathbf{Q}^t_c + \mathbf{Q}^t_m.
    \label{eq:Qini}
\end{equation}
This initialization scheme not only injects historical information into the queries but also facilitates their learning.\\
{\bf Memory Interaction.} To fully leverage the historical prior information stored in the memory bank, we design two structurally identical interaction modules, as illustrated in Figure.~\ref{overview}(d). The frame memory interaction module enhances the details of the current window using memory information, while the query memory interaction module learns temporal semantics from the memory. Together, they collaboratively mitigate the information truncation caused by non-overlapping windows.\\
{\bf Position-aware Interaction.} The Frame Memory Interaction module preliminarily embeds position-aware queries with historical motion information. To better localize the current gesture using historical context, we facilitate interaction between these queries and the current window frame features. The updated frame and query features are then used to update the memory bank. The query features are processed by detection and classification heads consisting of fully connected layers to predict micro gesture classes and their positions within the current window, achieving unified gesture detection and classification.

Overall, the learnable queries gain memorability through two mechanisms: interaction with frame-level memory and initialization using window-level memory. This enables them to continuously accumulate historical information during temporal reasoning, thereby assisting predictions in the current window.

\subsection{Training and Inference}
\label{ssec:4.4}
{\bf Training.} We use the standard bipartite matching loss~\cite{carion2020end} to supervise training. It finds an optimal matching between predicted queries and ground-truth gestures, ensuring each query corresponds to the most relevant instance. The loss consists of classification and position terms, jointly supervising each query's class and position.

The classification loss \( \mathcal{L}_{cls} \) adopts cross-entropy loss to measure the discrepancy between the predicted category probabilities of the queries and the GT labels. For each matched pair \((q_i, g_j)\) where \(q_i\) is a predicted query and \(g_j\) is a GT gesture. All unmatched queries are considered as background class. The classification loss is calculated as:
\begin{equation}
\mathcal{L}_{cls} 
= - \sum_{i} \log p_{q_i}(y_i),
\quad
y_i =
\begin{cases}
c_{j}, & m(i) = j,\\
c_{\varnothing}, & m(i) = \varnothing,
\end{cases}
\label{eq:Lcls_cases}
\end{equation}
where $m(i)$ denotes the index of the ground-truth gesture assigned to query $q_i$ under the optimal matching. If $m(i) = j$, the target label $y_i$ is set to the corresponding gesture category $c_j$; otherwise, $y_i$ is assigned to the background category $c_{\varnothing}$ for unmatched queries. $p_{q_i}(c)$ represents the predicted probability that query $q_i$ belongs to category $c$. The classification loss supervises both matched and unmatched queries, reducing false positive predictions.

The position loss \( \mathcal{L}_{pos} \) is designed to supervise the position parameters (center point coordinates \(x\) and width \(w\)) of the predicted gestures. It combines the L1 loss for the center coordinates and width and IOU loss, and is formulated as:
\begin{equation}
\begin{split}
\mathcal{L}_{pos} = \sum_{i,j} & \mathbbm{1}_{\{m(i)=j\}} (|x_{q_i} - x_{g_j}| + |w_{q_i} - w_{g_j}| \\
& + (1 - IoU(q_i, g_j)) ),
\end{split}
\label{eq:Lpos}
\end{equation}
where \(\mathbbm{1}_{\{m(i)=j\}} \) is an indicator function that is 1 if query \(q_i\) is matched to ground-truth \(g_j\) under the optimal matching \(m\). $x_{q_i}$ and $w_{q_i}$ are the predicted center coordinates and width of query $q_i$. $x_{g_j}$ and $w_{g_j}$ are the corresponding ground-truth values. $IoU(q_i, g_j)$ represents the intersection over union between the predicted gesture region of query $q_i$ and the ground-truth gesture region $g_j$.

Additionally, to better handle short and subtle micro gestures, we propose a query-based CTC loss that enforces temporal alignment between predicted gestures and ground truth. This loss reduces the network’s tendency to predict blank tokens for weak signals, improving its ability to recognize fast and continuous micro gestures. Finally, the overall loss of the network is defined as:\\
\begin{equation}
\mathcal{L} = \lambda_{cls}\mathcal{L}_{cls} + \lambda_{pos}\mathcal{L}_{pos} + \lambda_{q-CTC}\mathcal{L}_{q-CTC}.
    \label{eq:Lall}
\end{equation}
{\bf Inference.} During inference, the framework processes the input skeleton sequence with non-overlapping sliding windows. Position-aware queries, initialized from window-level memory, interact via cross-attention with frame-level memory and current features to capture local and historical information. These queries are passed to detection and classification heads to predict gesture categories and locations within the window. Low-confidence predictions are filtered out to produce the final results.
\section{Experiments}
\label{sec:exper}
\begin{table}[t]
    \centering
    \small
    \setlength\tabcolsep{0.9mm} 
    \renewcommand{\arraystretch}{0.9}
    \caption{Benchmark of SOTA methods with six metrics. Methods marked with * denote our re-implementations owing to the absence of open-source code.} 
    \begin{tabular}{lcccccc}
        \toprule
    \multirow{2}*{\textbf{Methods}}  
      & \multirow{2}*{\textbf{DR$\uparrow$}}  
      & \multirow{2}*{\textbf{FP$\downarrow$}}  
      & \multirow{2}*{\textbf{JI$\uparrow$}}  
      & \multirow{2}*{\textbf{NLD$\uparrow$}}  
      & \textbf{Infer.} & \textbf{Avg.} \\
    &&&&&\textbf{Time$\downarrow$} & \textbf{Delay$\downarrow$} \\
        \midrule
        \multicolumn{7}{c}{\textit{Offline-trained Sliding Window Methods}} \\
        \midrule
        ST-GCN~\cite{STGCN}                  & 75.1\% & 0.35 & 0.67 & 0.66 & 1.92 &7.84\\
        CTR-GCN~\cite{CTRGCN}                 & 76.4\% & 0.31 & 0.63 & 0.64 & 6.33 &8.03\\
        FR-Head~\cite{FRHead}                 & 78.1\% & 0.34  &  0.66  & 0.69  & 6.58 &8.11\\
        HD-GCN~\cite{HDGCN}                 & 77.8\% & 0.33 & 0.59 & 0.68 & 22.64 &8.57\\
        BlockGCN~\cite{zhou2024blockgcn}       &  78.3\% &  0.32 & 0.66 & 0.65 & 7.83 & 7.89\\
        HiOD~\cite{chang2025hierarchical}      &  81.2\% & 0.29 & 0.66 & 0.70 & 72.44&8.66\\
        \midrule
        \multicolumn{7}{c}{\textit{CTC-based Methods}} \\
        \midrule
        *STMG~\cite{kin2024stmg}               & 73.1\% & 0.36 & 0.59 & 0.72 & 2.89 & 8.28\\
        *HD-GCN+CTC             & 77.3\% & 0.38 & 0.64 & 0.71 & 22.64 &8.23\\
        \midrule
        \multicolumn{7}{c}{\textit{Boundary Supervision Methods}} \\
        \midrule
        OO-dMVMT~\cite{cunico2023oo}               & 79.1\% & 0.28 & 0.63 & 0.73 & 2.47 & 8.34 \\
        AG-MAE~\cite{ikne2025ag}           & 80.7\% & 0.28 & 0.65 & 0.72 & 2.36 & 8.25 \\
        *Det~\cite{kopuklu2019real}          & 66.4\% & 0.41 & 0.43 & 0.55 & 25.15 & 7.98\\
        *Bound.Reg.~\cite{HDGCN} & 81.6\% & 0.37 & 0.59 & 0.61 & 22.64 & 8.82\\
        \midrule
        \rowcolor{gray!20} \textbf{HMATr}          & \textbf{89.2\%} & \textbf{0.22} & \textbf{0.71} & \textbf{0.77} & \textbf{1.61} & \textbf{7.67} \\
        \bottomrule
    \end{tabular}
    \label{tab:sota}
\end{table}
\subsection{Experimental Setup}
\label{ssec:5.1}
{\bf Implementation Details.} We implement our method with PyTorch framework and perform all experiments on one RTX4090D GPU. We train our models using Adam with a weight decay of 0.0004. The batch size is set to 64 and the base learning rate is set to 0.001. The weights in the loss function are set as: $\lambda_{cls}=2$, $\lambda_{pos}=5$, $\lambda_{q-CTC}=0.2$. The number of learnable queries is set to 10, $L_f$ is set to 16, and $L_w$ is set to 3. The classification confidence threshold during online inference is set to 0.7.\\
{\bf Evaluation Metrics.} Following prior works ~\cite{cunico2023oo,shen2022gesture}, we employed four metrics to evaluate the performance of the model: Detection Rate (DR), False Positive Score (FP), Jaccard Index (JI), and Normalized Levenshtein Distance (NLD). We employed two metrics to evaluate the efficiency of the model: Inference Time (ms) on an RTX 4090D GPU and Average Delay (frames). More details about the metrics can be found in the supplementary materials.

\subsection{OMG-Bench Benchmark Evaluation}
\label{ssec:5.2}
{\bf Classification results.} We conduct experiments on OMG-Bench with three types of methods. \textbf{(1) Offline-trained Sliding Window Methods:} mainstream skeleton-based action or gesture recognition methods trained offline on pre-segmented datasets, followed by online inference using sliding windows. These methods determine gesture boundaries by thresholding the classification scores output from the classification heads. \textbf{(2) CTC-based Methods:} trained with the CTC loss~\cite{kin2024stmg}, detecting gesture boundaries by predicting blank tokens to determine the start and end of gestures. \textbf{(3) Boundary Supervision Methods:} incorporate boundary ground truth to assist classification in addition to category supervision. These mainly include two-stage approaches that first detect then classify (\textit{e.g.,} HD-GCN+Det~\cite{kopuklu2019real}), as well as methods introducing boundary-aware auxiliary losses (\textit{e.g.,} HD-GCN+Bound.Reg. and OO-dMVMT~\cite{cunico2023oo}). The results in Table \ref{tab:sota} show that our query-based method significantly outperforms others, with improvements of 7.6\%, 0.15, 0.12, and 0.16 on four evaluation metrics, respectively. This is mainly attributed to the learnable queries, which alleviate sensitivity to window size and stride, and fully encode the positional and semantic information of micro gestures during training.\\
{\bf Model efficiency.} As shown in Table \ref{tab:sota}, we compare the Average Delay (frames) and Inference Time (ms) on an RTX4090D GPU (batchsize=1). HMATr attains the lowest latency and fastest inference by combining (1) a non-overlapping sliding window, (2) hierarchical memory that preserves multi-level temporal context without re-extraction, and (3) position-aware learnable queries that unify detection and classification in one pass. Together, these choices make HMATr well‑suited for real‑time VR/AR micro gesture interaction.\\
\begin{table}[t]
    \centering
    \small
    \caption{Results on SHREC'22 and SHREC'21 benchmark.} 
    \begin{tabular}{lccc|cc}
        \toprule
        \multirow{2}*{\textbf{Methods}} 
            & \multicolumn{3}{c|}{\textbf{SHREC'22}} 
            & \multicolumn{2}{c}{\textbf{SHREC'21}} \\
        & \textbf{DR$\uparrow$} & \textbf{FP$\downarrow$} & \textbf{JI$\uparrow$} 
        & \textbf{DR$\uparrow$} & \textbf{FP$\downarrow$} \\
        \midrule
        DG-STA~\cite{chen2019construct} & 0.51 & 0.32 & 0.21 & 0.81 & 0.07 \\
        SeS-GCN~\cite{sampieri2022pose} & 0.60 & 0.16 & 0.53 & 0.75 & 0.12 \\
        PSUMNET~\cite{trivedi2022psumnet} & 0.62 & 0.24 & 0.52 & 0.64 & 0.22 \\
        MS-G3D~\cite{liu2020disentangling} & 0.68 & 0.21 & 0.57 & 0.69 & 0.25 \\
        DSTA~\cite{shi2020decoupled} & 0.73 & 0.24 & 0.61 & 0.81 & 0.08\\
        DDNet~\cite{DDNET} & 0.88 & 0.16 & 0.78 & 0.82 & 0.10 \\
        uDeepGRU~\cite{caputo2019shrec} & - & - & - & 0.85 & 0.10\\
        OO-dMVMT~\cite{cunico2023oo} & 0.92 & 0.09 & 0.85 & 0.88 & 0.05 \\
        \midrule
        \rowcolor{gray!20} \textbf{HMATr} & \textbf{0.95} & \textbf{0.08} & \textbf{0.88} & \textbf{0.92} & \textbf{0.04} \\
        \bottomrule
    \end{tabular}
    \label{tab:shrec}
\end{table}

\vspace{-5mm}  
\subsection{Results on Online Macro Gesture Datasets}
\label{ssec:5.3}
To evaluate the generalization of HMATr, we compare it with state-of-the-art methods on two additional online skeleton-based gesture recognition datasets, SHREC'22 and SHREC'21. Table \ref{tab:shrec} shows that HMATr achieves state-of-the-art performance, indicating that our approach applies not only to micro gesture recognition but also to conventional gesture recognition.\\

\vspace{-4mm}  
\subsection{Ablation Studies}
\label{ssec:5.4}
We conduct ablation studies on our OMG-Bench. In addition, we provide more results in supplementary materials.\\
{\bf Effectiveness of Each Component.} We designed three model variants to evaluate different components. (1) We removed the frame-level memory bank and replaced the cross-attention in Frame Memory Interaction with self-attention. (2) We removed the window-level memory bank and initialized the position-aware queries to zero. (3) We removed both the position-aware queries and the window-level memory bank, and obtained gesture predictions by averaging frame-level logits within each window. The first two variants were trained with the loss in Equation \ref{eq:Lall}, while the third used frame-level cross-entropy loss. As shown in Table \ref{tab:ab1}, the full model outperforms all three variants on all metrics. This shows that both the hierarchical memory bank and position-aware queries are effective, and that memory features further improve the ability of position-aware queries to detect and classify micro gestures.\\
\begin{table}[t]
    \centering
    \setlength\tabcolsep{0.8mm} 
    \renewcommand{\arraystretch}{0.9}
    \caption{Ablation study of different components. ‘FMB’ is frame-level memory bank. ‘WMB’ is window-level memory bank. ‘PQ’ is position-aware queries.}
    \begin{tabular}{lccccccc}
        \toprule
         & \textbf{FMB} & \textbf{WMB} & \textbf{PQ} & \textbf{DR$\uparrow$} & \textbf{FP$\downarrow$} & \textbf{JI$\uparrow$} & \textbf{NLD$\uparrow$} \\
        \midrule
        w/o FMB & \ding{55} & \ding{51} & \ding{51} & 86.1\% & 0.26 & 0.66 & 0.74 \\
        w/o WMB &\ding{51}& \ding{55} & \ding{51} & 87.0\% & 0.23 & 0.64 & 0.75 \\
        w/o PQ & \ding{51} & \ding{55} & \ding{55} & 82.4\% & 0.30 & 0.63 & 0.66 \\
        \rowcolor{gray!20} \textbf{HMATr} & \ding{51} &\ding{51} & \ding{51}& \textbf{89.2\%} & \textbf{0.22} & \textbf{0.71} & \textbf{0.77} \\
        \bottomrule
    \end{tabular}
    \label{tab:ab1}
\end{table}
{\bf Effectiveness of Interaction Modules.} To verify the effectiveness of the three interaction modules, we sequentially removed each module from the full model, as shown in Table \ref{tab:ab2}. Removing any module degraded performance, with the Position-aware Interaction module having the largest impact. This suggests that the learnable queries mainly extract gesture location and semantic information from current frame features, while the Frame/Query Memory Interaction module enhances frame and query features by leveraging historical information.\\
\begin{table}[t]
    \centering
    \renewcommand{\arraystretch}{0.9}
    \caption{Ablation study of different interaction modules. ‘PI’ is Position-aware Interaction module. ‘FMI’ and ‘QMI’ denote Frame and Query Memory Interaction modules, respectively.}
    \begin{tabular}{lcccc}
        \toprule
         & \textbf{DR$\uparrow$} &\textbf{FP$\downarrow$} & \textbf{JI$\uparrow$} & \textbf{NLD$\uparrow$} \\
        \midrule
        w/o PI          & 82.4\% & 0.30 & 0.59 & 0.68 \\
        w/o FMI         & 87.5\% & 0.28 & 0.69 & 0.72 \\
        w/o QMI         & 85.3\% & 0.25 & 0.63 & 0.70 \\
        w/o FMI + QMI   & 84.5\% & 0.27 & 0.64 & 0.75 \\
        w/o FMI + QMI + PI & 76.8\% & 0.35 & 0.62 & 0.69 \\
        \rowcolor{gray!20} \textbf{HMATr}           & \textbf{89.2\%} & \textbf{0.22} & \textbf{0.71} & \textbf{0.77} \\
        \bottomrule
    \end{tabular}
    \label{tab:ab2}
\end{table}
\begin{figure}[t]
	\centering
	\includegraphics[width=\linewidth]{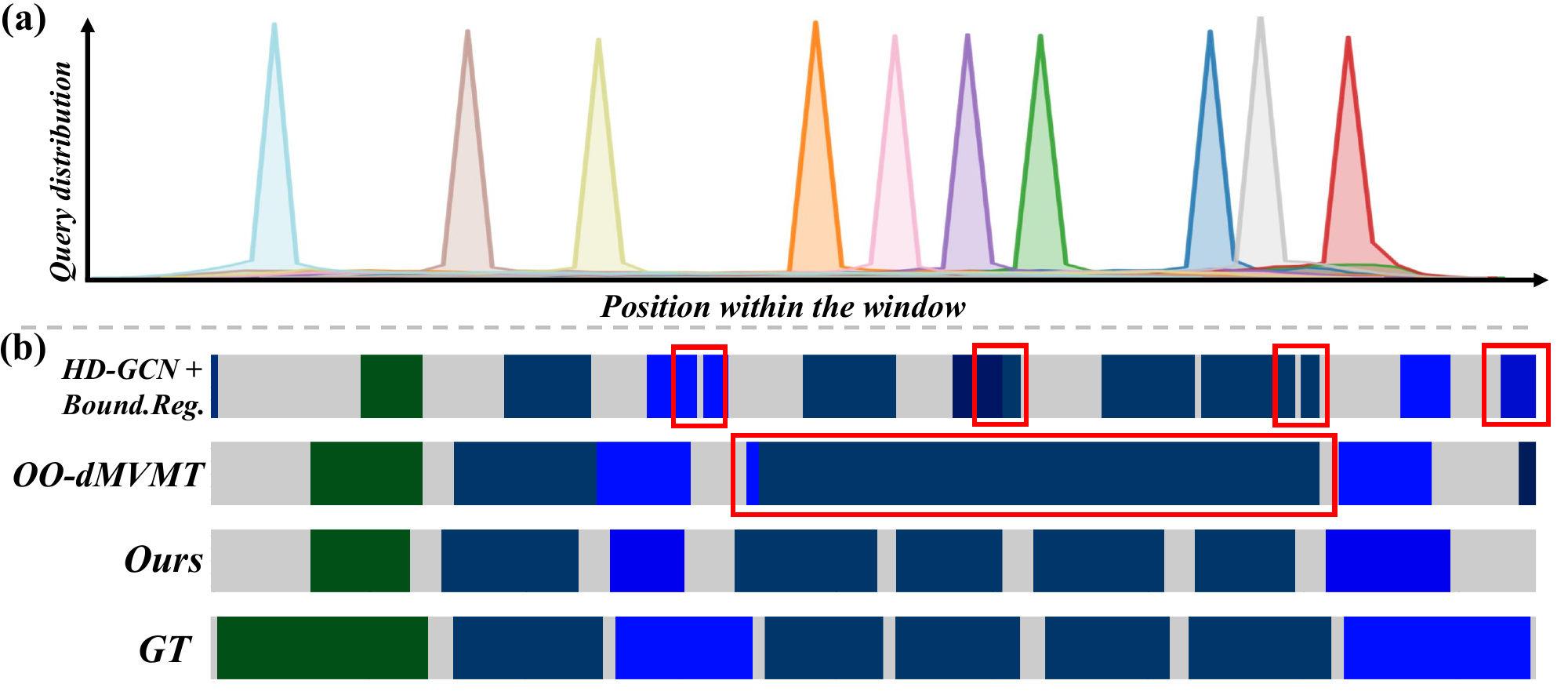}
	\caption{Visualization of (a) query distribution and (b) online recognition results of the gesture sequence.}
	\vspace{-4mm}  
	\label{vis}
\end{figure} 
\vspace{-3mm}  
\subsection{Qualitative Results}
\label{ssec:5.5}
{\bf Query Distribution.} In Figure \ref{vis}(a), we present the distribution of query relative positions within the window for all test samples. We observe that different queries tend to focus on different positions, thereby capturing diverse types of micro gesture features. This enhances the feature representation capability of HMATr.\\
{\bf Comparison of Gesture Segment Detection Results.} In Figure \ref{vis}(b), we visualize the micro gesture detection results of different methods (OO-dMVMT~\cite{cunico2023oo} and HD-GCN with boundary supervision~\cite{HDGCN}). The results demonstrate that our method can accurately identify the boundaries of consecutive same-class gestures, whereas the other two methods tend to either merge multiple consecutive same-class gestures or over-segment them. This highlights the superiority of our approach.
\section{Conclusions}
\label{sec:conclusions}
In this paper, we introduce OMG-Bench, the first large-scale public dataset for skeleton-based online micro gesture recognition. Its design captures rapid, continuous micro interactions with frame-level annotations. We also propose HMATr, a novel end-to-end framework using hierarchical memory and position-aware queries to unify detection and classification. Experimental results demonstrate that HMATr achieves SOTA performance with efficient non-overlapping window inference.

\section*{Acknowledgments}
This work was supported by the National Natural Science Foundation of China (Grant No.62332019, 62406039), the National Key Research and Development Program of China (Grant No.2023YFF1203900, 2023YFF1203903), the Beijing Nova Program (Grant No.20240484513).\\
{
    \small
    \bibliographystyle{ieeenat_fullname}
    \bibliography{main}
}

\clearpage
\setcounter{page}{1}
\maketitlesupplementary


In the supplementary material, we provide:

\begin{itemize}
\item[$\bullet $] more dataset details in Sec.\ref{sec:dataset},
\item[$\bullet $] more implementation details in Sec.\ref{sec:implementation},
\item[$\bullet $] more experiments in Sec.\ref{sec:exper}.
\item[$\bullet $] a video demo in Sec.\ref{sec:video}.
\end{itemize}

\section{More Dataset Details}
\label{sec:dataset}
\subsection{Class Definition}
\label{ssec:1.1}
\begin{figure}[t]
	\centering
	\includegraphics[width=\linewidth]{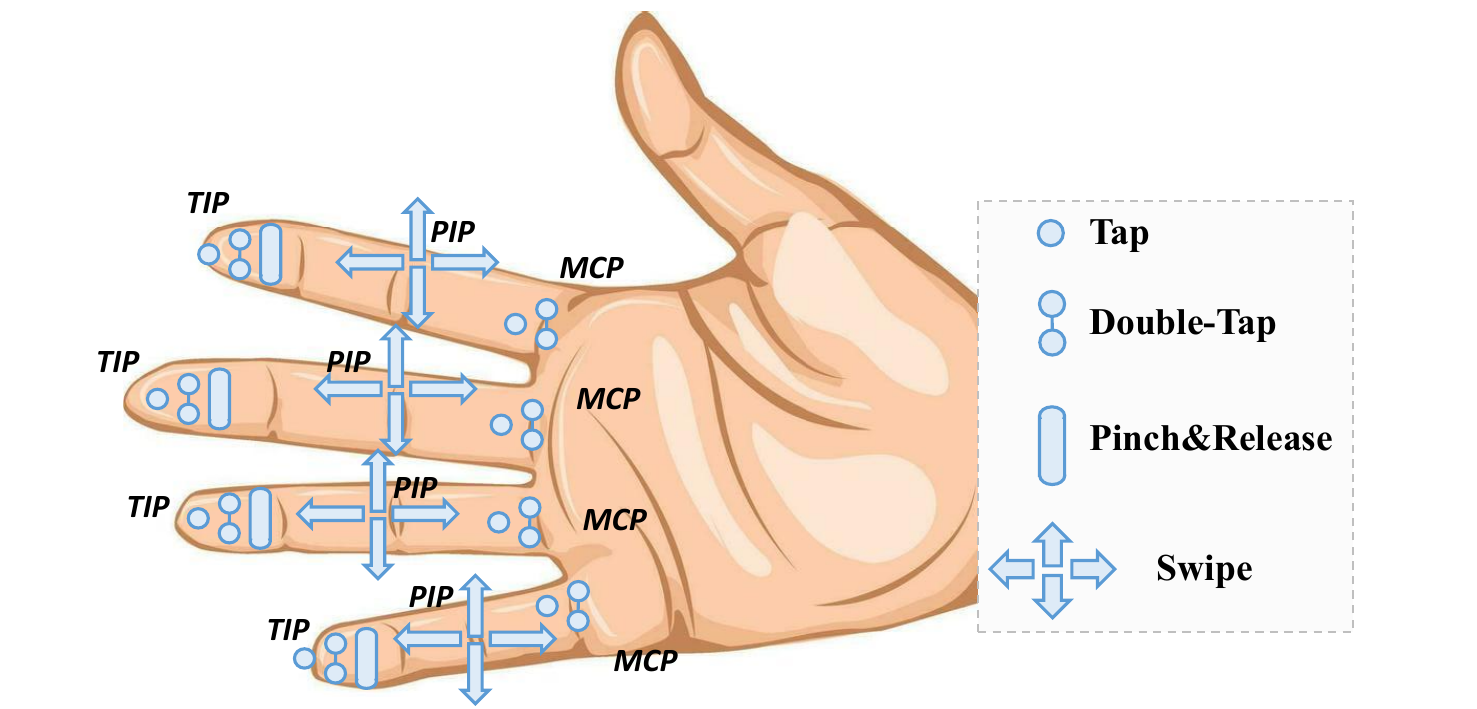}
	\caption{Types and interaction locations of all defined micro gestures. TIP, PIP, and MCP are anatomical terms for finger parts, referring to the finger tip, proximal interphalangeal joint, and metacarpophalangeal joint respectively.}
	\label{supp_hand}
\end{figure}

We define 40 common single-hand micro gesture classes, as illustrated in Table \ref{tab:gesture}. These gestures are all realized through interactions between the thumb and other fingers. The class differences are characterized by three dimensions: \textbf{(1) Interacting fingers}, i.e., the thumb interacting with the index, middle, ring, or little finger. \textbf{(2) Interaction types}, including typical actions such as single tap, double tap, slide, pinch, and release. \textbf{(3) Interaction locations}, referring to the specific contact areas between the thumb and finger—for example, taps occur at the TIP or MCP joint, while sliding covers a larger finger region. Figure \ref{supp_hand} illustrates the interaction locations and types of all micro gestures. Due to the compact spatial arrangement of hand joints and the small temporal differences between different gesture types (e.g., single tap and double tap), varying degrees of confusion arise in both spatial and temporal dimensions, which increase the challenges of online micro gesture recognition.
\begin{table*}[h]
	\centering
    \caption{All gesture classes and their corresponding descriptions defined in the OMG-Bench dataset.}
	\begin{tabular}{ccccl}
		\toprule
		{\textbf{Class id}}  & \textbf{Finger}  & \textbf{Location} & \textbf{Type} & \textbf{Class description} \\
		\midrule
            1 &Index &  TIP&Tap  &Thumb taps TIP of the index finger\\
            2 &Index & MCP &Tap  &  Thumb taps MCP of the index finger\\
            3  &Middle & TIP &Tap  &Thumb taps TIP of the middle finger  \\
            4 &Middle & MCP &Tap  &Thumb taps MCP of the middle finger  \\
            5 &Ring & TIP &Tap  & Thumb taps TIP of the ring finger \\
            6 &Ring & MCP &Tap  & Thumb taps MCP of the ring finger \\
            7 &Little & TIP &Tap  & Thumb taps TIP of the little finger \\
            8 &Little & MCP &Tap  & Thumb taps MCP of the little finger \\
            9  &Index & TIP &Double-Tap & Thumb double-taps TIP of the index finger \\
            10  &Index & MCP &Double-Tap &Thumb double-taps MCP of the index finger  \\
            11  &Middle & TIP &Double-Tap & Thumb double-taps TIP of the middle finger \\
            12  &Middle & MCP &Double-Tap & Thumb double-taps MCP of the middle finger \\
            13  &Ring & TIP &Double-Tap & Thumb double-taps TIP of the ring finger \\
            14  &Ring & MCP &Double-Tap & Thumb double-taps MCP of the ring finger \\
            15  &Little & TIP &Double-Tap &  Thumb double-taps TIP of the little finger\\
            16  &Little & MCP &Double-Tap &  Thumb double-taps MCP of the little finger\\
            17  &Index & TIP$\to$MCP &Swipe &  Thumb swipes right along the index finger\\
            18 &Index & MCP$\to$TIP &Swipe & Thumb swipes left along the index finger \\
            19 &Middle & TIP$\to$MCP &Swipe & Thumb swipes right along the middle finger \\
            20 &Middle &MCP$\to$TIP&Swipe &  Thumb swipes left along the middle finger\\
            21 &Ring & TIP$\to$MCP &Swipe & Thumb swipes right along the ring finger \\
            22 &Ring & MCP$\to$TIP &Swipe & Thumb swipes left along the ring finger \\
            23 &Little & TIP$\to$MCP &Swipe & Thumb swipes right along the little finger \\
            24 &Little & MCP$\to$TIP &Swipe & Thumb swipes left along the little finger \\
            25 &Index & PIP &Swipe & Thumb swipes up/forward along the index finger \\
            26 &Index & PIP &Swipe & Thumb swipes down/backward along the index finger \\
            27 &Middle & PIP &Swipe &  Thumb swipes up/forward along the middle finger\\
            28 &Middle & PIP &Swipe & Thumb swipes down/backward along the middle finger \\
            29 &Ring & PIP &Swipe & Thumb swipes up/forward along the ring finger \\
            30 &Ring & PIP &Swipe & Thumb swipes down/backward along the ring finger \\
            31 &Little & PIP &Swipe & Thumb swipes up/forward along the little finger \\
            32 &Little & PIP &Swipe & Thumb swipes down/backward along the little finger \\
            33 &Index & TIP &Pinch\&Release& Thumb and index finger pinch \\
            34  &Index & TIP &Pinch\&Release & Thumb and index finger release \\
            35  &Middle & TIP &Pinch\&Release & Thumb and middle finger pinch \\
            36  &Middle & TIP &Pinch\&Release & Thumb and middle finger release \\
            37  &Ring & TIP &Pinch\&Release & Thumb and ring finger pinch \\
            38  &Ring & TIP &Pinch\&Release & Thumb and ring finger release \\
            39  &Little & TIP &Pinch\&Release & Thumb and little finger pinch \\
            40  &Little & TIP &Pinch\&Release & Thumb and little finger release \\
		\bottomrule
	\end{tabular}
	
	\label{tab:gesture}
\end{table*} 

\subsection{Data Collection and Annotation}
\label{ssec:1.2}
{\bf Interactive Tasks for Data Collection.} All subjects collected micro gesture sequences by performing randomly generated interactive tasks. Before the data collection began, each subject was informed of the data collection requirements and provided written informed consent for academic use. To ensure data quality, all subjects underwent standardized training before collection to familiarize themselves with the gesture standards. Sequences with non-standard motions or subject anomalies were immediately flagged and repeated.

\begin{figure}[t]
	\centering
	\includegraphics[width=\linewidth]{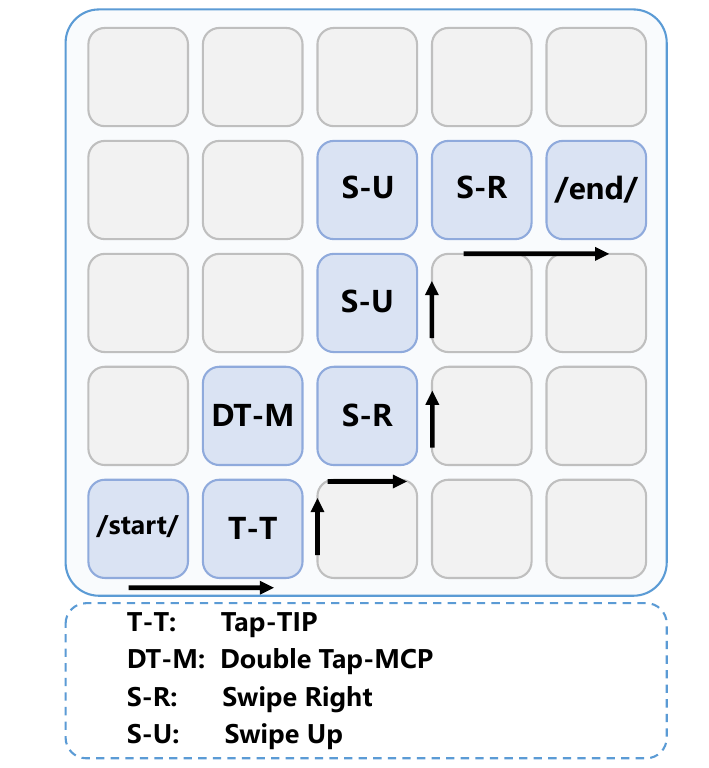}
	\caption{Illustration of the interactive task.}
	\label{supp_game}
\end{figure}

The interactive task was designed as a gesture-triggered “grid-walking” game, as illustrated in Figure \ref{supp_game}. Subjects were required to sequentially move a virtual block along a predefined grid sequence, where each grid cell required the completion of a specific gesture to proceed to the next. A gesture sequence collection was completed once all grid cells were traversed. Each execution path on the grid was randomly generated, and the generated gesture classes varied accordingly. Typically, a sequence contained 8 to 16 micro gestures to ensure diversity within the gesture sequences. To align with human intuitive interaction, the interactive task tends to generate gestures that are consistent with the actual movement of the block during micro gesture generation. For example, a right-swipe gesture controlled the block to move right, while single and double taps controlled turning. As a result, the dataset contains some consecutively performed gestures of the same class, which better reflects realistic VR/AR interaction scenarios.

To enable subjects to trigger interactive tasks without an online micro gesture recognition model, we designed a threshold-based heuristic micro gesture recognition algorithm. This heuristic algorithm identifies gestures by measuring the distances between the thumb and specific joints of other fingers. Our primary objectives were (1) to provide subjects with real-time interactive feedback to ensure natural gesture execution, and (2) to determine the start and end of gestures via the heuristic algorithm, enabling automated gesture frame labeling. Therefore, the classification accuracy of the heuristic algorithm was not critical; rather, its accuracy in detecting gesture occurrences was essential. For example, since the system knows that the target gesture is a thumb sliding along the index finger, it can determine the start and end frames by measuring the distance between the thumb tip and either the index fingertip or the base of the index finger, without considering the other fingers. This greatly reduces the difficulty of heuristic recognition, thereby maximizing the chance that each gesture triggers the virtual module to move to the next grid cell. Finally, we retain only the gesture boundaries predicted by the heuristic algorithm, while the gesture labels are directly assigned from the predefined sequence.\\
{\bf Human Annotation.} Five professional volunteers familiar with gesture interaction and motion annotation were recruited to inspect and correct any erroneous automatic labels, ensuring label accuracy.  Specifically, experts review skeleton-overlaid RGB frames to rectify class labels and boundary inaccuracies. If the gestures performed by subjects differed from predefined sequence, predefined labels were modified to match. If initial boundaries differ significantly from the actual motion, they are manually re-annotated. Compared to traditional labor-intensive manual annotation methods, our approach significantly improved accuracy and avoided uncertainty caused by human bias.\\
{\bf Inter-Annotator Consistency.} We invited five experts to jointly annotate 5\% of the data and computed the pairwise boundary IoU and boundary variance among them. The average boundary IoU was 0.81, and the average boundary variance was 2.13 frames, indicating high annotation consistency. Furthermore, after retraining with random boundary noise applied to the ground truth (±5 frames, p = 0.5), HMATr achieved a DR of 89.1\% (a drop of 0.1\%), while OO-dMVMT achieved a DR of 78.7\% (a drop of 0.4\%). These results demonstrate that annotation noise within 5 frames has a negligible impact on training.\\
{\bf Multi-view Hand Pose Automatic Annotation Algorithm.} OMG-Bench follows the standard 21-joint skeleton annotation scheme \cite{zhang20163d}, which includes the fingertip and three phalangeal joint centers  for each finger, along with the wrist joint. However, annotating large-scale hand pose datasets is both time-consuming and labor-intensive. Therefore, we employed a multi-view self-supervised hand pose estimation method to achieve automatic hand skeletal acquisition. A self-supervised keypoint detector is trained on depth images using the Dual-Branch Self-Boosting Framework (DSF) \cite{ren2022dual}, and applied to unlabeled multi-view sequences following the Cross-View Fusion Network strategy \cite{ren2022mining}. This automatic stage yields accurate and consistent 3D joint coordinates and MANO parameters under most conditions. We tested this method on the manually annotated held-out evaluation set, and the error is 2.78 mm.

\subsection{Dataset Attribute Metrics}
\label{ssec:1.3}
In Table \ref{tab:dataset2}, we use four metrics to measure the differences between OMG-Bench and SHREC’21 \cite{caputo2021shrec} and SHREC’22 \cite{emporio2022shrec}.\\
\textbullet \quad \textbf{Same-Class Continuous Gesture Percentage (SCCGP)}: The proportion of continuously performed gestures belonging to the same class relative to the total number of samples. OMG-Bench contains more consecutive gestures of the same class, better reflecting real-world VR/AR interaction scenarios. For example, when users browse web pages or select applications from VR virtual menus, they often perform consecutive downward swipe gestures to turn pages. At the same time, these consecutive same-class gestures are prone to being recognized as a single gesture, increasing boundary ambiguity in online gesture recognition.\\
\textbullet \quad \textbf{Mean Gesture Interval (MGI)}: The average time interval between consecutive gesture samples. The gesture intervals in OMG-Bench are relatively shorter, better reflecting the natural gesture interaction frequency of humans in VR/AR environments, but they also further increase boundary ambiguity in online gesture recognition.\\
\textbullet \quad \textbf{Mean Gesture Duration (MGD)}: The average duration of all gesture samples. This metric indicates that most gestures in OMG-Bench have significantly shorter durations compared to the other two datasets. Such a short response time characteristic of micro gestures increases the challenge of online recognition.\\
\textbullet \quad \textbf{Normalized Mean Joint Displacement (NMJD)}: The average joint displacement over all sequences and joints, normalized by the distance between the wrist joint and the root of the middle finger. This suggests that, compared to the other two datasets, OMG-Bench exhibits more subtle motion characteristics, making it challenging to precisely determine the position and class of micro gestures from fine-grained movements.\\
\begin{table}[t]
    \centering 
    \caption{Statistical Comparison of OMG-Bench with SHREC'21/22.}
    \setlength\tabcolsep{1.2mm} 
    \begin{tabular}{lcccccccc}
        \toprule
        & \textbf{SCCGP}$\uparrow$ & \textbf{MGI}$\downarrow$ & \textbf{MGD}$\downarrow$ & \textbf{NMJD}$\downarrow$\\
        \midrule
        SHREC'21~\cite{caputo2021shrec}& 0.29\% & 12.48s & 2.60s & 158.38 \\
        SHREC'22~\cite{emporio2022shrec}& 0.09\% & 2.86s & 1.19s & 128.73\\
        \textbf{OMG-Bench} & \textbf{27.60\%} & \textbf{0.22s} & \textbf{0.57s} & \textbf{8.95}\\
        \bottomrule
    \end{tabular}
    
    \label{tab:dataset2}
\end{table}

\section{More Implementation Details}
\label{sec:implementation}
\subsection{Training Details}
\label{ssec:2.1}
We implement our method with PyTorch framework and perform all experiments on one RTX4090D GPU. We train our models using Adam with a weight decay of 0.0004. The batch size is set to 64 and the base learning rate is set to 0.001. We adopted a cosine learning rate scheduling strategy with a 5-epoch warm-up phase. The loss weights were set through grid search as follows: $\lambda_{cls}=2$, $\lambda_{pos}=5$, $\lambda_{q-CTC}=0.2$. The number of learnable queries is set to 10, the length of the frame-level memory bank is set to 16, and the length of the window-level memory bank is set to 3. The Frame Memory Interaction, Query Memory Interaction, and Position-aware Interaction modules were each configured with 1 layer. The number of attention heads in both the self-attention and cross-attention modules is set to 4. The classification confidence threshold during online inference is set to 0.7. The feature extraction backbone comprises 3 ST-GCN layers~\cite{STGCN} with channel configurations 3→64, 64→128, and 128→256.

Since the length of each gesture sequence varies, we divided each sequence into shorter segments of 128 frames for temporal training. The sliding step for segmentation was set to 64 frames to ensure effective utilization of all data. Within each short segment, we applied a non-overlapping sliding window scheme with a window size of 16 frames and a stride of 16 frames to train the proposed HMATr. As a result, each short segment contained 8 continuous windows.

For other baselines, we follow the training paradigm commonly adopted in typical skeleton-based human action recognition methods ~\cite{CTRGCN, HDGCN, chang2025hierarchical, WDCE} as well as previous online gesture recognition approaches ~\cite{cunico2023oo} for all compared methods. Specifically, we first perform offline training on the pre-segmented dataset, and then adopt a sliding-window strategy during online inference, with a window size of 16, a stride of 1, and a Logits score threshold of 0.8.

\subsection{Query-based CTC loss}
\label{ssec:2.2}
We employed an auxiliary loss $\mathcal{L}_{q-CTC}$ based on Connectionist Temporal Classification (CTC) \cite{graves2006connectionist} alongside the Hungarian matching loss. Specifically, we supervised the classification results of matched queries within each 128-frame short segment using the standard CTC loss. This leverages CTC’s ability to predict blank tokens to suppress outputs from irrelevant queries, thereby further enhancing the queries’ capability to perceive gesture boundaries. Related ablation studies are presented in Sec.\ref{ssec:3.1}.

\subsection{Benchmark Evaluation Metrics}
\label{ssec:2.3}
We employed six metrics to evaluate the performance of the model.\\
\textbullet \quad \textbf{Detection Rate (DR)}: The ratio between the number of correctly detected gestures and the total number of gestures in the input sequences. A gesture is considered correctly detected if it has a temporal intersection with the ground truth greater than 50\% of the true interval, does not last more than twice the real duration, and has the same label. The gestures predicted by the recognizer but not corresponding to ground truth ones are defined as false positives.\\
\textbullet \quad \textbf{False Positive Score (FP)}: Defined as the ratio between the number of false positives and the total number of gestures.\\
\textbullet \quad \textbf{Jaccard Index (JI)}: The average relative overlap between the ground truth and the predicted labels for the input sequences. It is used in many continuous classification tasks, but it does not evaluate the ability to avoid multiple activations for a single gesture or small noisy activations.\\
\textbullet \quad \textbf{Normalized Levenshtein Distance (NLD)}: Defined as the average normalized edit distance between the predicted gesture sequence and the ground truth sequence. It measures the minimum number of insertions, deletions, and substitutions required to transform the predicted sequence into the ground truth, normalized by the length of the ground truth. NLD evaluates the overall sequence-level similarity and penalizes recognition errors including missed, spurious, and misclassified gestures. It can be expressed as:
\begin{equation}
    \text{NLD} = 1 - \frac{\text{levenshtein}(y_{\text{predict}}, y_{\text{true}})}{\text{length}(y_{\text{true}})},
\label{eq:Xframe}
\end{equation}
where $y_{\text{predict}}$ and $y_{\text{true}}$ are the predicted and true list of labels of the gestures respectively.\\
\textbullet \quad \textbf{Inference Time}: We perform model inference on an RTX 4090D GPU and measure the inference time (in milliseconds) with a batch size of 1. This metric serves as the most direct measure of the efficiency of online gesture recognition models.\\
\textbullet \quad \textbf{Average Delay}: Average Delay is defined as the difference, measured in frames, between the actual gesture end frame and the last frame reported by the algorithm for predicting the gesture start frame.

\begin{table}[t]
    \centering
    \caption{Ablation study of different memory bank lengths.}
    \begin{tabular}{cccccc}
        \toprule
          \textbf{Frame} & \textbf{Window} & \textbf{DR$\uparrow$} & \textbf{FP$\downarrow$} & \textbf{JI$\uparrow$} & \textbf{NLD$\uparrow$} \\
        \midrule
         1& 3 & 85.9\% & 0.28 &0.63 & 0.70\\
         8& 3 & 87.2\% &0.24 &0.68& 0.75\\
         \textbf{16}&\textbf{3} & \textbf{89.2\%} & \textbf{0.22} & \textbf{0.71} & 0.77 \\
         24& 3 & 88.9\% & 0.22&0.66& \textbf{0.79}\\
         16& 1 &87.6\% & 0.23 & 0.66&0.74 \\
         16& 2 &87.5\% & 0.22 & 0.68& 0.75 \\
         16& 4 &88.2\% & 0.25&0.69 &0.75 \\
        \bottomrule
    \end{tabular}
    
    \label{tab:aab1}
\end{table}

\section{More Experiments}
\label{sec:exper}
\subsection{Ablation Studies}
\label{ssec:3.1}
{\bf Effect of Memory Bank Length.} We investigate the impact of memory bank length on online micro gesture recognition by fixing the length of one memory bank while varying the length of the other. Results in Table \ref{tab:aab1} show that the best performance is achieved when the frame-level memory bank length is 16 and the window-level memory bank length is 3. We found that neither a memory bank that is too short nor one that is too long yields the best performance. This is because when the memory bank is too short, the model obtains limited historical temporal information from it. Conversely, when the memory bank is too long, the historical information may contain multiple completed independent gestures, which can interfere with the current gesture prediction.\\
\begin{table}[t]
    \centering
    \caption{Ablation study of the number of queries.}
    \begin{tabular}{ccccc}
        \toprule
           \textbf{Number of Queries} & \textbf{DR$\uparrow$} & \textbf{FP$\downarrow$} & \textbf{JI$\uparrow$} & \textbf{NLD$\uparrow$} \\
        \midrule
        3& 85.6\% & \textbf{0.21} & 0.65 & 0.76\\
        5& 86.2\% & 0.23 & 0.67 & 0.75\\
        \textbf{10}&  \textbf{89.2\%} & 0.22 & \textbf{0.71} & \textbf{0.77}\\
        20& 88.5\% & 0.26 & 0.66 &0.74 \\
        \bottomrule
    \end{tabular}
    
    \label{tab:aab2}
\end{table}
\begin{table}[t]
    \centering
    \caption{Ablation study of Query-CTC loss.}
    \begin{tabular}{ccccc}
        \toprule
          \textbf{Loss}  & \textbf{DR$\uparrow$} & \textbf{FP$\downarrow$} & \textbf{JI$\uparrow$} & \textbf{NLD$\uparrow$} \\
        \midrule
        w/o $\mathcal{L}_{q-CTC}$ & 88.8\% & 0.25 & 0.69 & 0.73 \\
        \textbf{Ours}&  \textbf{89.2\%} & \textbf{0.22} & \textbf{0.71} & \textbf{0.77}\\
        \bottomrule
    \end{tabular}
    
    \label{tab:aab3}
\end{table}
{\bf Effect of the Number of Queries.} We conduct ablation studies to investigate the effect of the number of queries on online micro gesture recognition. Experiments are performed with the number of queries set to 3, 5, 10, and 20. Results in Table \ref{tab:aab2} show that the overall performance is best when the number of queries is 10. When the number of queries is too small, their diversity in representing positional and semantic features is limited, making it difficult to capture the complexity and variability of micro gestures. Meanwhile, their positional awareness is relatively sparse, failing to represent multiple possible locations. Conversely, when the number of queries is too large, irrelevant queries may be activated, resulting in redundant gesture detections.\\
{\bf Effectiveness of Query-CTC Loss.} We validate the effectiveness of the Query-CTC loss through ablation experiments. Table \ref{tab:aab3} presents the results of experiments without the Query-CTC loss. We observe that removing the Query-CTC loss significantly affects the normalized Levenshtein distance (NLD), while other metrics show minor changes. This demonstrates that the Query-CTC loss primarily optimizes the network learning from the perspective of the overall sequence. The Query-CTC loss can suppress the activation of irrelevant queries to some extent, preventing redundant gesture outputs. Additionally, it injects global semantic information into the position-aware queries at the sequence level, enhancing the feature representation capability of the queries.\\
{\bf Effect of Global Memory Embedding Strategy.} To investigate how to effectively exploit the global historical information contained in the window-level memory, we explore three global memory embedding strategies for the queries: (1) Zero initialization: no global history is embedded; position-aware queries are directly zero-initialized. (2) Cross-attention: position-aware queries attend to the window-level memory features via cross-attention to inject global historical information. (3) Memory initialization: position-aware queries are initialized with the mean-pooled feature of the window-level memory bank (Section 4.3 in the main text). As shown in Table \ref{tab:aab4}, our memory initialization achieves the best performance for embedding historical information. Moreover, compared to cross-attention, memory initialization introduces no additional learnable parameters and is therefore more efficient.

\begin{table}[t]
    \centering
    \caption{Ablation study of global memory embedding strategy.}
    \setlength\tabcolsep{1.6mm} 
    \begin{tabular}{ccccc}
        \toprule
           \textbf{Strategy} & \textbf{DR$\uparrow$} & \textbf{FP$\downarrow$} & \textbf{JI$\uparrow$} & \textbf{NLD$\uparrow$} \\
        \midrule
        Cross-Att. & 89.0\% & \textbf{0.20} & 0.70 & 0.76 \\
        Zero init. & 88.5\% & 0.24 &  0.68 & 0.72 \\
        \textbf{Memory init. (Ours)}&  \textbf{89.2\%} & {0.22} & \textbf{0.71} & \textbf{0.77}\\
        \bottomrule
    \end{tabular}
    \label{tab:aab4}
\end{table}

\subsection{More Quantitative Experiments}
\label{ssec:3.2}
{\bf Effect of Skeleton Accuracy in the Dataset.} To evaluate the effect of skeletal data accuracy in the dataset on the performance of HMATr, we conducted inference-stage experiments on test sets with varying levels of skeletal data precision. Specifically, we first generate hand skeleton sequences for the entire test set using different hand pose estimation algorithms (e.g., WiLoR~\cite{potamias2025wilor}, HaMeR~\cite{pavlakos2024reconstructing}, MMVI-single~\cite{ren2022mining}), relying solely on monocular RGB-D data from camera 0. We then directly perform inference on these skeleton sequences --- produced by various pose estimation algorithms --- using HMATr trained on the OMG-Bench training set. In other words, the training data consist of skeletons obtained via multi-view fusion from the original OMG-Bench dataset, whereas the testing data consist of skeletons estimated from single-view inputs by different algorithms.

As shown in Table \ref{tab:quant1}, the accuracy drop of HMATr varies across test sets with different skeletal data qualities, which is primarily constrained by the quality of skeletal estimations in the test set. MMVI-single~\cite{ren2022mining}, a single-view variant of our multi-view fusion algorithm for skeletal data generation, produces skeletal estimations of relatively higher quality compared to the RGB-based methods WiLoR~\cite{potamias2025wilor} and HaMeR~\cite{pavlakos2024reconstructing}, thereby achieving comparatively higher testing accuracy. Moreover, the limited accuracy degradation observed in Table \ref{tab:quant1} also reflects, to some extent, the generalization capability of HMATr, indicating its applicability to skeletal data estimated by different algorithms.\\
\begin{table}[t]
    \centering
    \caption{Results of inference using HMATr trained on the original training set on test sets derived from different Hand Pose Estimation (HPE) algorithms.}
    \small
    \setlength\tabcolsep{1.6mm} 
    \begin{tabular}{cccccc}
        \toprule
         \textbf{HPE Method}  & \textbf{Modality} & \textbf{DR$\uparrow$} & \textbf{FP$\downarrow$} & \textbf{JI$\uparrow$} & \textbf{NLD$\uparrow$} \\
        \midrule
        WiLoR~\cite{potamias2025wilor} & RGB &  73.6\% &  0.15 & 0.64 &  0.68 \\
        HaMeR~\cite{pavlakos2024reconstructing} & RGB &  77.3\% &  0.14 &  0.67 & 0.73 \\
        MMVI-single~\cite{ren2022mining} & Depth &  83.7\% & 0.20  & 0.73  & 0.77 \\
        \bottomrule
    \end{tabular}
    \label{tab:quant1}
\end{table}
\begin{figure}[t]
	\centering
	\includegraphics[width=\linewidth]{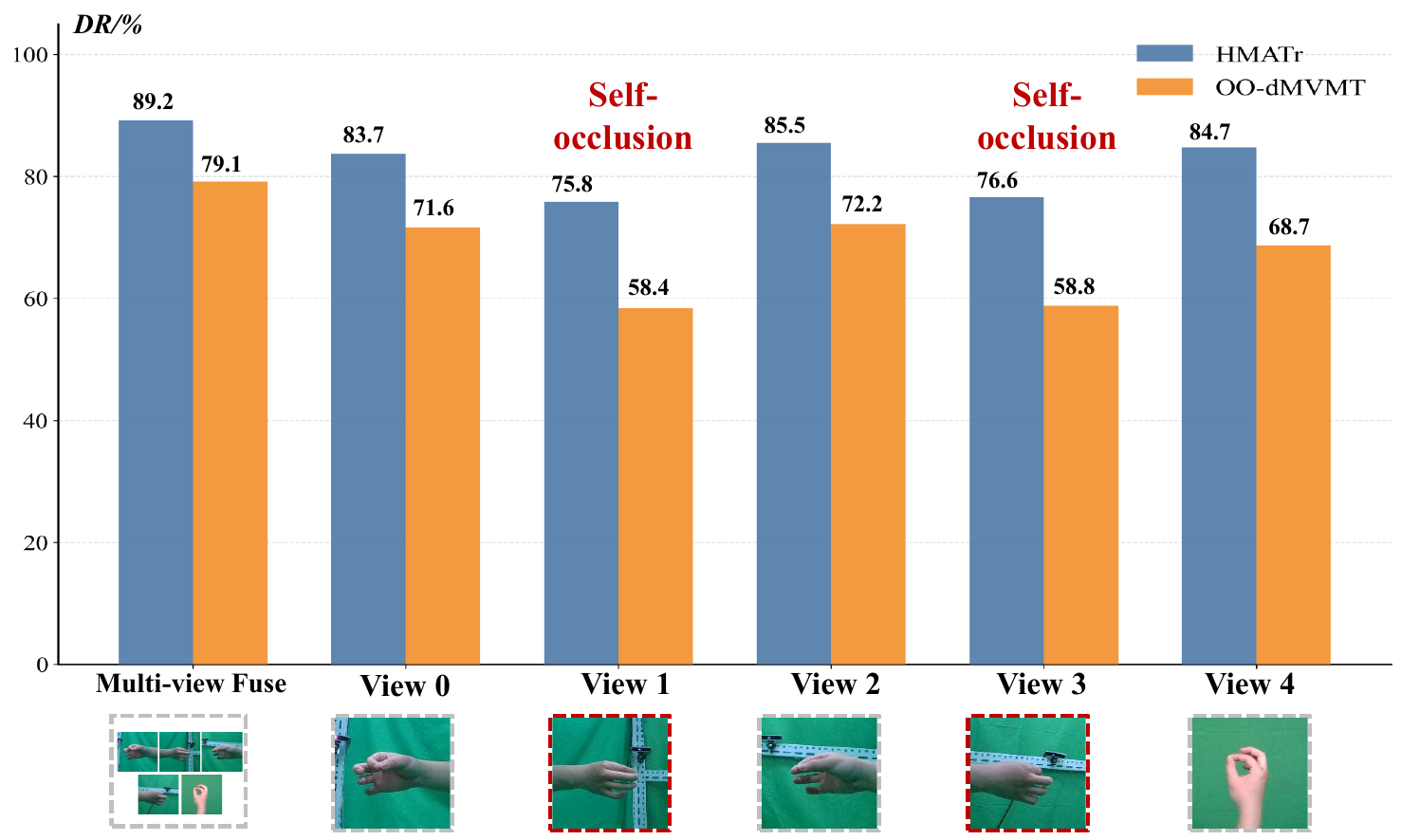}
	\caption{Inference results of HMATr and OO-dMVMT trained on OMG-Bench under single-view settings. The skeleton data for each view are obtained using a single-view hand pose estimation algorithm to simulate the inaccuracies of skeleton estimation in real-world applications.}
	\label{supp_view}
\end{figure}
{\bf Single-view generalization.} During training, we apply skeleton rotation and translation normalization to improve the robustness of our method to large-amplitude motions and arbitrary viewpoints. To better reflect real-world conditions, we perform inference using skeletons estimated from each individual view. Figure \ref{supp_view} presents the test results under different viewpoints. We observe that even in extreme views such as View1/3, where severe self-occlusion occurs, HMATr still maintains a satisfactory gesture detection rate, demonstrating its robustness. Meanwhile, the highest recognition accuracy is achieved under multi-view fusion, highlighting the necessity of acquiring high-quality skeleton data through multiple views.\\

\vspace{-4mm}
\subsection{Confusion Analysis}
\label{ssec:3.3}
\begin{figure}[t]
	\centering
	\includegraphics[width=\linewidth]{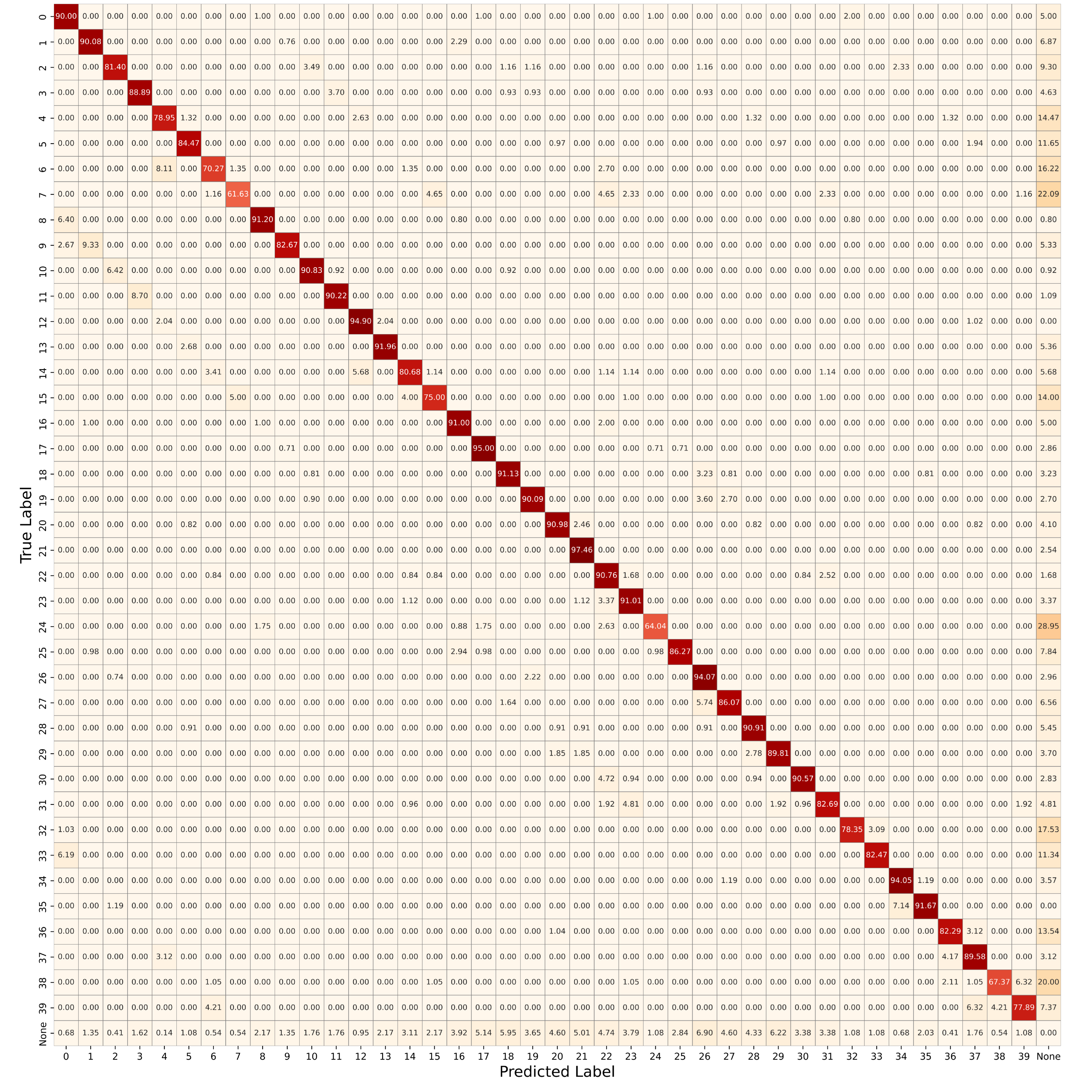}
	\caption{The confusion matrix of HMATr on OMG-Bench.}
	\vspace{-3mm}  
	\label{supp_confusion}
\end{figure}

Figure \ref{supp_confusion} illustrates the confusion patterns of our HMATr for online micro-gesture recognition on the OMG-Bench dataset. We identify the top three most confused class pairs as class 9/class 1, class 6/class 4, and class 11/class 3. Analysis of these three challenging pairs reveals two main difficulties: (1) temporal confusion: rapid double-tap gestures of the index and middle fingers are often misclassified as single taps, with error rates of 9.86\% and 8.79\%, respectively, due to the short temporal intervals in continuous inputs; (2) anatomical confusion: tapping with the little finger is easily confused with tapping with the ring finger (error rate: 9.68\%), owing to the strong coupled motion between these two adjacent fingers.

\section{Video Demo}
\label{sec:video}

In our project page, we provide a video to further illustrate visualization details in the OMG-Bench dataset and the practical deployment of the proposed HMATr.

The video showcases samples from the 40 micro gesture classes included in OMG-Bench, covering four common interaction types: Swipe, Tap, Double-Tap, and Pinch\&Release. The inherent properties of OMG-Bench --- rapid dynamics, subtle motion, high inter-class similarity, and frequent same-class continuity --- pose new challenges for online micro gesture recognition.

Additionally, we deploy HMATr, trained on the OMG-Bench dataset, onto the Quest Pro headset to evaluate practical micro gesture interaction. We directly leverage the headset’s built-in high-precision hand pose estimation to acquire skeletal data in real time. The generalization primarily stems from skeleton rotation and translation normalization applied during training and from HMATr’s hierarchical memory mechanism with position-aware queries. Specifically, the hierarchical memory leverages historical context to effectively smooth inter-frame jitter and occlusion-induced noise prevalent in monocular estimation, while the attention-based position-aware queries focus on relative motion patterns rather than absolute coordinates, enabling implicit semantic alignment across heterogeneous skeleton structures and thereby ensuring robust cross-algorithm inference. Moreover, the high-precision skeleton data in OMG-Bench further contributes to this generalization. The video demonstrates that the proposed online micro gesture recognition algorithm supports interaction via multiple micro gestures while ensuring real-time performance and low latency.

\end{document}